\documentclass{article}

\usepackage{neurips_data_2024}

\usepackage[utf8]{inputenc} 
\usepackage[T1]{fontenc}    
\usepackage{hyperref}       
\usepackage{url}            
\usepackage{booktabs}       
\usepackage{amsfonts}       
\usepackage{nicefrac}       
\usepackage{microtype}      
\usepackage{xcolor}         
\usepackage{graphicx}

\title{PufferLib: Making Reinforcement Learning \\ Libraries and Environments Play Nice}

\author{Joseph Suárez \\
  \texttt{jsuarez@puffer.ai}
}

\begin{document}

\maketitle

\begin{abstract}

You have an environment, a model, and a reinforcement learning library that are designed to work together but don’t. PufferLib makes them play nice. The library provides one-line environment wrappers that eliminate common compatibility problems and fast vectorization to accelerate training. With PufferLib, you can use familiar libraries like CleanRL and SB3 to scale from classic benchmarks like Atari and Procgen to complex simulators like NetHack and Neural MMO. We release pip packages and prebuilt images with dependencies for dozens of environments. All of our code is free and open-source software under the MIT license, complete with baselines, documentation, and support at pufferai.github.io.

\end{abstract}

\section{Background and Introduction}

Continued progress in reinforcement learning (RL) requires training on increasingly sophisticated environments. Nearly all of modern RL tooling was written with Atari in mind: arcade games with flat image observations, discrete actions, and a single agent. As a result, many of the most interesting environments are incompatible with standard tools. These oversimplifications have led to more pernicious implicit assumptions as well that make interesting environments even less practical to work with. For example, by assuming that each environment instance will run at roughly the same speed as all the others, environments with deeply branching logic become too slow to use. And dependency management is so poor that even installing environments can be a hassle.

\begin{figure}
    \centering
    \includegraphics[width=\linewidth]{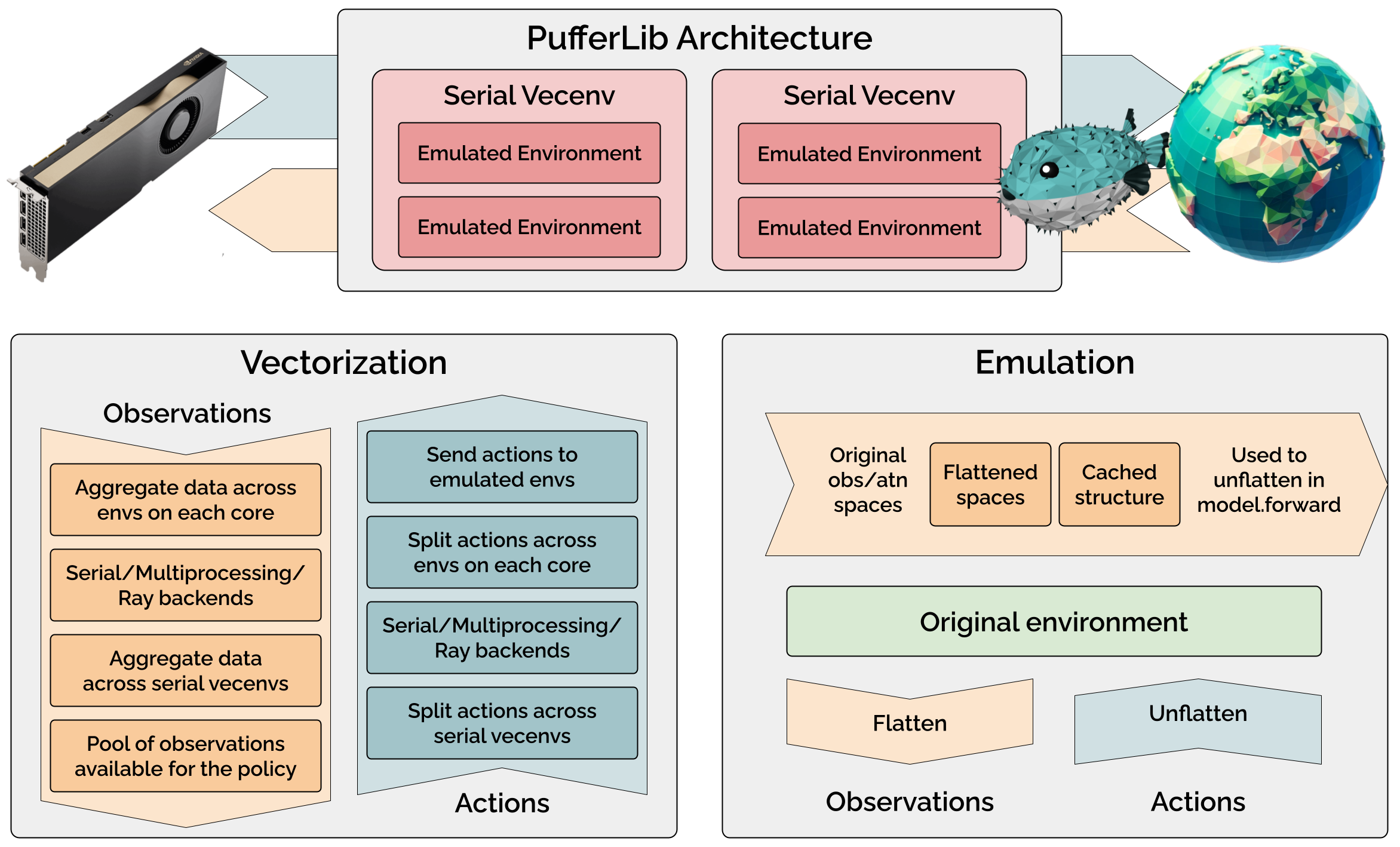}
    \caption{The PufferLib system architecture for broad environment compatibility and fast vectorization. Each core simulates one or several environments. Vectorization aggregates data from several processes and distributes actions across them. Each environment is wrapped in PufferLib's core emulation layer, which ensures flat data representations.}
    \label{fig:header}
\end{figure}

PufferLib solves all of these problems. The key insight that started this project is that it is possible wrap environments to \textit{look} like Atari from the perspective of learning libraries, without any loss of generality. This means we can use all of the standard tools, exactly as they are, with nearly any environment. We then built high-performance vectorization for distributed simulation, bindings for dozens of popular environments, and additional tools for developers. Our main contributions are:

\begin{enumerate}
    \item One-line wrappers that make complex environments like Nethack, Neural MMO, Griddly, etc. compatible with any RL library that supports standard Gymnasium/PettingZoo formats.
    \item Drop-in vectorization for simulating environments in parallel. Most environments will see at least a 30\% speed boost and 50\%-3x with pooling. This is a broadly compatible contribution applicable to nearly all game environments.
    \item Open-source demos with bindings for a dozen+ common environments. The entire training stack for the most complex environment is <2,000 lines of code.
\end{enumerate}

The system architecture is shown in Figure \ref{fig:header}. Unlike most heavy RL frameworks, PufferLib provides a light-weight set of tools that simply improve performance and compatibility with other RL libraries. PufferLib integrates with all common environment formats:

\textbf{Gym/Gymnasium} \citep{DBLP:journals/corr/BrockmanCPSSTZ16, towers_gymnasium_2023} These libraries define the API for the vast majority of environments used in research and provide associated tools, such as the \texttt{spaces} module for defining data shapes and \texttt{vector} for vectorization. An important piece of historical context: OpenAI originally published Gym in 2016, and nearly all non-DeepMind environments from 2016-2022 used it. Maintenance was taken over by an independent open source group in 2021, but since they did not have full control of the repository required for maintenance, they forked it as Gymnasium in 2022. Thus, when we say that PufferLib supports Gym and Gymnasium, note that this involves a nontrivial chain of dependency management and compatibility patches that would normally fall to the user.

\textbf{PettingZoo:} \citep{DBLP:journals/corr/abs-2009-14471} A multiagent analog to Gym/Gymnasium. Even though it was developed independently, the PettingZoo API followed the Gym/Gymnasium schism and has similar versioning and dependency issues as a result. PufferLib handles these as well.

\textbf{DM Env:} \citep{dm_env2019} DeepMind's counterpart to Gym. Both provide comparable APIs, and there is no real reason to use one over the other, but most modern research has standardized on Gymnasium. PufferLib converts to the Gymnasium API by default for this reason. 

PufferLib is not another SB3 or CleanRL because it is not a learning library. It includes only one first-party PPO implementation, which is itself adapted from CleanRL, and even that only exists for specific testing reasons. Instead, PufferLib makes all of the existing libraries work the way they are supposed to. To those outside of RL, this may seem like an incredibly low bar, but fighting with basic infrastructure issues for weeks is a nearly universal experience among RL researchers and practitioners. We include a list of related libraries that may include some utilities, though they primarily target learning itself.

\begin{enumerate}
    \item \textbf{CleanRL:} \citep{DBLP:journals/corr/abs-2111-08819} Single-file implementations with research-friendly features. Best-in-class for simplicity and ease of prototyping new algorithms. No modularity by design.
    \item{\textbf{Stable Baselines 3 (SB3):} \citep{JMLR:v22:20-1364} Reliable implementations for a wide range of algorithms. Best-in-class for newcomers applying RL algorithms to a supported problem. Harder to modify deeply.}
    \item{\textbf{Sample Factory 2.0:} \citep{petrenko2020sf} High-throughput reinforcement learning. Best in class for pure performance. Hardest to modify, fewest algorithmic options.}
    \item{\textbf{Tianshou:} \citep{tianshou} Another widely used RL library. Comparable in design to SB3 with similar tradeoffs.}
    \item{\textbf{TorchRL:} \citep{bou2023torchrl} Official PyTorch RL library. This is a new project that does not yet have a specific identity.}
\end{enumerate}

Our evaluation of strengths and weaknesses is subjective, but we expect it to resonate with the RL community. Note that this list omits libraries focused on control and robotics environments, most of which are simulated on the GPU or are in the process of being ported to the GPU. PufferLib currently focuses on tooling for CPU environments. Running environments on GPU enables massively faster training at the price of constraining the types of environments that can be used for research. As a result, we consider these parallel areas of tooling, each of which is unlikely to replace the other.

In summary, PufferLib sands the rough edges off of environments and learning libraries to sit in the middle. Gym/Gymnasium/DM Env define how data enters and exits the environment. PufferLib formats that data in a way that learning libraries expect and provides performance improvements along the way.

\section{PufferTank}

\textbf{Problem:} Setting up a development environment for RL is difficult and time consuming. Extensive system package dependencies, slow source builds, and versioning obstacles prevent otherwise useful environments from attaining broader adoption.

\textbf{Solution:} PufferTank is a Docker container with everything you need set up correctly and tricky versioning issues handled. The current version has CUDA 12.1, PyTorch 2.4, Python 3.11, dependencies for dozens of environments, and convenient development features such as a rendering passthrough for WSL. We are aware that containerized development is not ubiquitous in AI and thoroughly considered lighter weight options like a Conda environment. Unfortunately, the dependencies for many environments take hours to build and contain a mix of system and python packages. Our prebuilt images allow you to get up and running in minutes and to reset to the base installation instantly if anything goes wrong. Most IDEs, including VSCode, have plugins for seamless development in both local and remote containers. For minimalists, NeoVim with SuperMaven code completion is preinstalled.

Note that \textbf{you can use PufferLib without PufferTank}. PufferLib itself is a PyPI package (\texttt{pip install pufferlib}) -- PufferTank exists because \textit{specific common environments} have tricky dependencies, not PufferLib itself.

\section{PufferLib}

\subsection{Emulation}

\textbf{Problem:} Learning libraries make strong assumptions that are incompatible with cognitively interesting and efficient environments. This is partially a historical problem and partially an engineering limitation. Most learning libraries were designed with Atari in mind, which is single-agent with flat tensor (image) observations and single discrete actions. The most interesting environments usually do not adhere to these constraints. For example, NetHack is rendered in ASCII with additional information contained in a panel at the bottom of the console. There is no sensible way to represent this information as a flat tensor, so the NetHack Learning Environment \citep{DBLP:journals/corr/abs-2006-13760} is forced to expose it as a dictionary of arrays with different shapes. Even when flat representations are possible, they are not always optimal. For example, Neural MMO \citep{nmmo_neurips} provides isometric rendering, but the training API instead exposes local state data for each of a variable number of agents because it is 1000x faster, enabling users to train on a desktop instead of a supercomputer.

\textbf{Solution:} PufferLib provides one-line wrappers that make simple learning libraries work with complex environments. It does so by flattening observations to tensors and actions to a single multidiscrete variable. This means that, from the perspective of the learning library, the environment \textit{looks} like Atari, thereby \textit{emulating} a simpler environment. PufferLib provides a function to undo this operation, which you can call in the first line of your model's forward pass. This means that there is \textbf{no loss of generality}.

Normally, to use an environment like NetHack, you would first have to write a single-purpose environment wrapper to do something similar to what PufferLib does more generally. Since this requires packing and unpacking arbitrary data, it is an error prone operation that is difficult to test. PufferLib does all of this for you with an efficient implementation tested against dozens of real and mocked environments.

PufferLib's approach is so simple as to seem obvious, but there are many seemingly reasonable ways to accomplish the same task that do not work. For example, several reinforcement learning libraries attempt to natively support hierarchical observation and action data with no flattening. This dramatically complexifies the code base and also prevents multiple important optimizations later in the data pipeline, most notably during vectorization. As shown in Figure 1, emulation overhead is negligible for environments slower than several thousand steps per second per core.

PufferLib's emulation layer also handles other niche but important compatibility issues. For example, it will perform shape checks on the first batch of data. This catches nearly all user errors but does not add any overhead, since the checks are only performed at startup. In multiagent environments, PufferLib ensures that observations and actions are returned in a canonical sorted order. If the environment has a variable number of agents, PufferLib will pad observations to maintain fixed size data buffers. These are all common sources of difficult to diagnose bugs.

\subsection{Environments}

PufferLib currently provides bindings for atari, procgen, nethack, neural mmo, magent, minigrid, minihack, crafter, griddly, pokemon, and more. These are not the only supported environments, just the ones that we have manually tested for dependency and versioning quirks. An updated list is available at pufferai.github.io.

\textbf{Problem:} Many widely used environments have unresolved versioning and API compatibility issues with Gym/Gymnasium. This is in addition to the system dependencies resolved by PufferTank.

\textbf{Solution:} PufferLib provides known-good bindings for dozens of popular environments. These include Gym/Gymnasium conversion, the standard emulation wrapper, and sometimes additional fixes for specific quirks. There are installation options for each package, e.g. \texttt{pip install pufferlib[atari]} or \texttt{pip install pufferlib[nethack]}. PufferLib provides additional insurance against poor dependency management -- for example, we pin versions for sub-packages that are know to commonly ship breaking changes. There is also a \texttt{pufferlib[common]} option, which installs the broadest set of mutually compatible environments. This is included with PufferTank by default, so most users should have an out-of-the-box development experience. It was also one of the motivations for building PufferTank in the first place, since some of the \texttt{common} environments require additional system packages. Note that \textbf{pufferlib does not have a registry} by design, so there are \textbf{no additional requirements for custom environments}.

\subsection{Vectorization}

Vectorization is the process of simulating many environments, usually on different cores, which requires aggregating observations and distributing actions. Learning libraries either use Gym/Gymnasium's builtin vectorization or ship their own.

\textbf{Problem:} Existing vectorization methods are slow and provide limited or no support for complex environments with e.g. structured observations, multiple agents, variable population size, etc.

\textbf{Solution:} PufferLib implements fast and broadly compatible vectorization from scratch. We provide serial, multiprocessing, and Ray backends with the same API. The greatest focus is on the multiprocessing backend. \textbf{All of the above combined are implemented in only a few hundred lines.} For additional details, the code is truly quite simple to read.

\textit{A hard assumption on PufferLib emulation.} Earlier, we said that other libraries attempting to implement native structured data processing would cause issues. This is one such instance. Gymnasium and SB3, the two most popular vectorization implementations, both attempt native support. The Gymnasium implementation misses several crucial opportunities for optimization as a result. The SB3 implementation simply flattens observations, without giving the user any way to unflatten them. For some reason, it does this on the main process and with a rather inefficient implementation. By comparison, PufferLib's implementation is shorter, faster, and more flexible. The features described below would be quite difficult to implement otherwise.

\textit{Native multiagent support.} Most vectorization implementations, including PettingZoo's, bolt this on with unoptimized wrappers. Both SB3 and Gymnasium have made clear that there will never be official multiagent support.

\textit{A Python implementation of EnvPool} \citep{weng2022envpool}. Standard vectorization simulates $M$ environments in parallel and requires waiting on all $M$ before returning observations. PufferLib can instead retrieve $N<<M$ observations. This has two important implications. First, by setting $M = 2N$, simulation becomes approximately double-buffered. This means that CPU cores are processing half of the environments while the GPU computes actions for the other half. Second, by setting $M \gg 2N$, the model no longer has to wait on the slowest environments before returning a batch of observations. This feature is especially important in complex environments, which tend to have more branching logic paths. At time of writing, this is the only existing Python implementation of EnvPool, as opposed to the original EnvPool implementation that supports select C++ environments.

\textit{Multiple environments per worker.} PufferLib allows you to specify the number of environments and the number of workers separately. When a worker is responsible for multiple environments, it efficiently stacks data returned by each sub-environment in preallocated arrays without performing any extra copies. This feature is important when running many more environments than your machine has cores, as it avoids clogging the system with small processes.

\textit{Shared memory for data communication.} We load \texttt{observation}s, \texttt{reward}s, \texttt{terminal}s, \texttt{truncated}s, and \texttt{action}s signals into large shared arrays. We use pipes, which are up to 10x faster than queues due to a Python quirk, for communicating \texttt{info}s. Empty \texttt{infos} are pruned, and we provide wrappers to aggregate them over episodes. As a result, only one step per episode requires any inter-process communication. By comparison, SB3 does not have shared memory. Gymnasium provides a slower shared memory implementation that attempts to handle structured data natively, requiring multiple small copy operations and additional Python logic.

\textit{Shared flags for signaling.} Worker processes busy-wait on an unlocked shared array flag to detect when actions are ready and update the flag after computing observations. This almost completely eliminates inter-process communication overhead. Pipes are only used when an environment returns non-empty \texttt{info}s, which will be once per episode when using our wrappers.

\textit{Zero-copy batching.} For environments with large observations, we provide a setting to load batches of data directly from shared memory by waiting on a contiguous subset of worker process indices. Other settings require one copy. A naive implementation not using shared memory would require two or three copies.

\textit{Four separately optimized code paths.} For fast environments, main process overhead has to be optimized to within a few microseconds. Even operations like manipulating process IDs in a list can result in noticeable performance drops. We identify and separately optimize four common workload cases. In the synchronous case, environments are split evenly across cores and loading into a single batch in shared memory with no extra copy operations. In the fully asynchronous case, data is taken from the first workers to finish, requiring a single copy operation to load the batch into contiguous memory. There is a special case of the latter where each batch is simulated on a single worker, so it can be loaded without additional copies. There is also the above zero-copy case, which is roughly equivalent to a circular buffer of batches.

\textit{An autotune utility.} Obtaining the best configuration for your environment and hardware requires testing all four code paths. We provide an utility that benchmarks valid vectorization settings.

\subsection{Models}

PufferLib provides an optional model format that splits the normal PyTorch \texttt{forward} function into separate \texttt{encode} and \texttt{decode} functions. This allows PufferLib to sandwich an LSTM between the computation of hidden state and actions. We apply this operation as a wrapper, meaning that LSTM support becomes optional and configurable per-experiment, without having to write two models. Users are free to disregard this feature and use other frameworks for LSTM support or implement it manually. However, our users have found this feature important, since LSTM state reshaping operations are one of the most common sources of difficult to diagnose bugs.

Pufferlib includes a few baseline models such as the original NatureCNN and ResNet models from DQN \citep{DBLP:journals/corr/MnihKSGAWR13} Impala \citep{DBLP:journals/corr/abs-1802-01561}, respectively. There is also a default architecture which defines an MLP sized to the flat observation and action spaces. This is useful for small test environments. \textbf{All base models directly subclass torch.nn.Module.} There is no additional model layer required by PufferLib.

\section{First-party Environments: Puffer Ocean}

\textbf{Problem:} Discovering breaking changes to algorithm implementations, which often result from seemingly innocuous tweaks, requires training several environments for multiple hours each and the spare hardware to do so.

\textbf{Solution:} Puffer Ocean is a suite of environments that are trivial with correct implementations and impossible with specific common bugs. Each environment trains in under a minute on a single CPU core, and the entire suite can be trained, tracked, and saved on WandB over a quick coffee break. The current set of environments are:

\begin{itemize}
    \item \textbf{Squared:} Agent starts at the center of a square grid. Targets are placed on the perimeter of the grid. Reward is 1 minus the L-inf distance to the closest target. This means that reward varies from -1 to 1. Reward is not given for targets that have already been hit.
    \item \textbf{Password:} Guess the password, which is a static binary string. The policy has to not determinize before it happens to get the reward, and it also has to latch onto the reward within a few instances of getting it.
    \item \textbf{Stochastic:} The optimal policy is to play action 0 $p$ percent of the time and action 1 $(1 - p)$ percent of the time. This is a test of whether the algorithm can learn a nonuniform stochastic policy.
    \item \textbf{Memory:} Repeat the observed sequence after a delay. It is randomly generated upon every reset. The sequence is presented one digit at a time, followed by a string of 0.
    \item \textbf{Multiagent:} Agent 1 must pick action 0 and Agent 2 must pick action 1.
    \item \textbf{Spaces:} A simple environment with hierarchical observation and action spaces. Obtaining maximal score requires taking into account all subspaces.
    \item \textbf{Bandit:} Simulates a classic multiarmed bandit problem. 
    
\end{itemize}

To be absolutely clear: \textbf{we never want to see scores on Ocean reported in a comparative baseline}. This is a sanity check only. Our PPO implementation solves each environment (score $> 0.9$) in roughly 30k interactions with a single set of barely tuned hyperparameters.

\section{Performance}

PufferLib's emulation layer typically adds a few tens of microseconds to simulation time. As shown in Table 1, the overhead is negligible for environments slower than a few thousand steps per second. Emulation works by inferring a numpy structured array datatype from the environment's Gym/Gymnasium observation and action spaces. This is an analog to C structs that provides an efficient numpy interface over structured data in contiguous memory. Conveniently, we can use structured arrays as flat bytes, as is required for efficient vectorization, or with dict-like accessors, as is required by the model and environment. This critical piece of code is Cythonized and tested to be faster than a half dozen implementations from earlier in development, including efforts to write it in C and Rust. Suffice that our implementation is, at the least, probably faster than the naive one-off script that users would have to write for any specific environment otherwise.

\begin{table}
    \centering
    \begin{tabular}{c|cccc}
         \textbf{Environment} &  \textbf{SPS} &  \textbf{\% Reset} &  \textbf{\% Step STD} & \textbf{\% Overhead} \\
         \hline
         Neural MMO & 2400 & 68 & 59 & 7.2 \\
         NetHack & 29k & 1.1 & 106 & 17 \\
         MiniHack & 11k & 2.1 & 28 & 4.9 \\
         Pokemon Red & 700 & 0.00 & 43 & 0.08 \\ 
         Cartpole & 270k & 18 & 37 & 13 \\
         Ocean Squared & 240k & 55 & 53 & 14 \\
         Procgen Bigfish & 25k & 0.36 & 14 & 2.5 \\
         Atari Breakout & 1.2k & 54 & 4.3 & 0.16 \\
         Nethack & 39k & 0.63 & 45.3 & 64 \\
         Crafter & 320 & 80 & 26 & 0.04 \\
         Minigrid & 16k & 4.5 & 8.1 & 2.7 \\
    \end{tabular}
    \caption{Updated single-core throughput and emulation overhead of various environments, evaluated on the desktop. Steps per second (SPS) is timed with emulation. Some environments have slow resets or high per-step variance.}
    \label{tab:updated_label_with_new_data_and_overhead}
\end{table}

\begin{table}
\centering
\begin{tabular}{c|cccc}
\textbf{Environment} & \textbf{Pufferlib (D/L)} & \textbf{Puffer Pool (D/L)} & \textbf{Gymnasium (D/L)} & \textbf{SB3 (D/L)} \\
\hline
Neural MMO & 13k / 4.5k  & 19k / 7.2k & - / - & - / - \\
Nethack & 96k / 18k & - / - & 7k / 7k & 6k / - \\
Minihack & 55k / 8k & - / - & 11k / 7k & 12k / 6k \\
Pokemon Red & 5k / 650 & 7.2k / 830 & 5k / 960 & 4.7k / 1k \\
Cartpole & 460k / 110k & 3M / 200k & 82k / 14k & 100k / 11k \\
Ocean Squared & 490k / 110k & 4M / 190k & 116k / 12k & 99k / 12k \\
Procgen Bigfish & 33k / 6.2k & 150k / 9.6k & 41k / 5k & 30k / 4.8k \\
Atari Breakout & 11.8k / 2k & 25.6k / 3k & 4.8k / 2.5k & 3.3k / 2.3k \\
Crafter & 360 / 170 & 2.8k / 450 & 340 / 135 & 350 / 203 \\
Minigrid & 151k / 16k & 210k / 20k & 54k / 10k & 44k / 10k \\
\end{tabular}
\caption{Vectorized throughput of PufferLib (with and without EnvPool), SB3, and Gymnasium. Evaluated on the desktop (D) and laptop (L).}
\label{tab:combined_vectorized_throughput}
\end{table}

PufferLib's vectorization is faster than both the Gymnasium and SB3 implementations in almost all cases, even without our EnvPool feature enabled, which provides most of the speedup. \textbf{The following comparisons are a worst case scenario for PufferLib} because they also rely on PufferLib's emulation. The fairer comparison would be to time PufferLib's emulation + vectorization vs. other implementations without PufferLib emulation. This would trigger inefficient code paths, such as Gymnasium's structured shared memory processing and SB3' main-thread flattening. However, several of the test environments simply would not work without PufferLib's emulation. By comparison, we had never explicitly tested any of these environments with Gymnasium/SB3 vectorization, and they all worked on the first try.

We benchmark a variety of real environments with different step times, step time variances, and observation/action spaces. We use two separate machines for testing. The first machine is a commercial desktop with a 24-core i9-14900k processor and an RTX 4090 GPU. The second machine is a laptop with a 6-core i7-10750H processor and an RTX 3070 GPU. Table 2 reports the results of these experiments.

The Gymnasium and SB3 multiprocessing implementations experience significant scaling degradation above 1000 synchronizations per second per core. Instead of clogging the system with small processes, PufferLib provides an optimized implementation for running multiple environments/core. This allows it to scale even to environments that run at 100k+ steps/second.

The other major source of performance improvement is PufferLib's EnvPool implementation. This is another place where our benchmarks are unfair to our own implementation: the tests are run without a model in a loop. A major benefit of EnvPool is allowing the environments to continue computing observations while the policy is computing actions. This can drive GPU idle time to 0. 

Even in this worst case scenario, we generally obtain at least a 30-40\% performance improvement with EnvPool. There are important cases where the improvement can be much larger. For example, Crafter is 6x faster with Puffer Pool. The reason for this is that Crafter has especially long reset times and high step time variance. You may have also noticed that PufferLib scales better on the desktop (D) than on the laptop (L). The chipset used has 8 \textit{performance cores} and 16 slower \textit{efficiency cores}. This is an increasingly common design in high-end Intel chips. SB3 and Gymansium are bottlenecked by the slowest environment and the slowest core, while PufferLib will retrieve observations from the first environments to finish processing. 

\section{First-party Training with Clean PuffeRL}

By design, PufferLib does not include a library of learning algorithm implementations and we have no plans to develop one. Instead, we provide integrations with CleanRL and SB3, with more integrations to be added based on user demand. With that said, CleanRL is designed to be modified by users, and we do maintain one heavily customized version of CleanRL's PPO \citep{DBLP:journals/corr/SchulmanWDRK17} implementation for testing and baselines. It has been expanded to allow separate training and evaluation, model saving and checkpointing, faster LSTM support, better logging and WandB integration, asynchronous environment simulation, and additional features for multiagent learning. We include a runner file with a CLI for all included PufferLib environments, clean YAML configs, and integration with WandB for tracking, baselines, and hyperparameter tuning.

\section{Proof of Impact}

PufferLib has already been widely used. We highlight two major successes that were powered by Puffer. Our emulation features made the NeurIPS 2023 Neural MMO competition possible. No learning library could handle Neural MMO 2.0 \citep{nmmo_neurips} natively, and PufferLib enabled participants to train competent policies in as little as 8 hours on a single desktop. For comparison, this is an environment that can take weeks to get up and running, despite good dependency management, purely because existing vectorization and learning libraries are not designed for environments of this complexity (many agents, variable population size, structured observations and actions, etc). This competition attracted 200 participants and far fewer complaints from participants about the complexity of the baseline than in previous years.

The second major project is RL for Pokemon Red, originally by Peter Whidden. Shortly after it was released in a viral YouTube video, we began helping port the project to PufferLib, onboarding interested contributors and providing computational resources. These results are not yet published, but the model has improved severalfold over the original project. The logging, visualization, and performance enhancements provided by PufferLib were instrumental in this success, and the current code for this is also open source. Training runs at 7000 steps per second with Clean PuffeRL on a single desktop, which is 2-3x the performance of the original SB3 project. This corresponds to roughly 3000x realtime simulation with an aggressive 24 frameskip.

\section{Limitations}

PufferLib does not yet support continuous action spaces. This is a relatively straightforward feature planned for within the next few minor updates. PufferLib does not yet have integrations with every major learning library. We are starting with CleanRL and SB3 and expanding based on user demand. PufferLib is missing some of the newer Gymnasium spaces. Few environments use these, and integrations are planned based on user demand.

There are a few edge cases in vectorization where PufferLib's synchronous multiprocessing is slower than Gymnasium/SB3. This is a quirk of OS process switching. We will either add a fallback option or find a workaround in the next update.

While we provide our integration with Neural MMO as evidence of our library's efficacy in complex multiagent domains, there are simply no good multiagent vectorization implementations to compare against. We fought with deprecated PettingZoo wrappers for a while, but they did not work with Neural MMO. This alone should be a good indicator of PufferLib's impact.

\section{Conclusion}

PufferLib provides fast and flexible tools for reinforcement learning. The majority of new users will benefit immediately from improved performance and a smoother development experience. However, the real value of PufferLib is the ease with which researchers can move their work to more complex and interesting environments, which have traditionally been difficult to work with. We hope that PufferLib will allow the field to explore new ideas hitherto constrained by the simplicity of readily available environments.

\bibliography{neurips_data_2024}
\bibliographystyle{abbrvnat}

\end{document}